\documentclass[pdflatex,sn-mathphys-num]{sn-jnl}% Math and Physical Sciences Numbered Reference Style
\usepackage{a4}%
\usepackage{graphicx}%
\usepackage{multirow}%
\usepackage{amsmath,amssymb,amsfonts}%
\usepackage{amsthm}%
\usepackage{mathrsfs}%
\usepackage[title]{appendix}%
\usepackage{xcolor}%
\usepackage{textcomp}%
\usepackage{manyfoot}%
\usepackage{booktabs}%
\usepackage{algorithm}%
\usepackage{algorithmicx}%
\usepackage{algpseudocode}%
\usepackage{listings}%
\theoremstyle{thmstyleone}%
\theoremstyle{thmstyletwo}%

\theoremstyle{thmstylethree}%

\begin{document}

\title[Article Title]{A Recursive Pyramidal Algorithm for Solving the Image Registration Problem}


\author[]{
    \fnm{Stefan}
    \sur{Dirnstorfer}
}
\affil[]{\orgname{testup.io}}


\abstract{
    The problem of image registration is finding a transformation that aligns two images,
    such that the corresponding points are in the same location.
    This paper introduces a simple, end-to-end trainable algorithm that is implementable in
    a few lines of Python code. The approach is shown to work with very little training data and training time, while
    achieving accurate results in some settings. An example application to stereo vision was trained from 74 images on a 19x15 input window.
    With just a dozen lines of Python code this algorithm excels in brevity and may serve as a good start in
    related scenarios with limitations to training data, training time or code complexity.
}




\maketitle

\section{Introduction}\label{sec1}

The image registration problem occurs when two images are taken from one scene with
differences in view point, timing and potentially other transformations.
The problem is to find a mapping between a point in one image onto a corresponding
point in the other. Ideally, one would like to trace all particles that are responsible for a visual impression
on each image. Because images are discretely sampled, this correspondence must be reconstructed from reasonable
assumptions of how the items in the scene could have transformed, including possible distortions, occlusions and
independent movements.

Practical applications of image registration include
stereo vision, optical flow and motion detection. It is also an important preparatory step to many
visual comparison tasks that are must be tolerant to changes in view points and structural distortions.
Early algorithms in this field used feature detection and an inference
mechanism for likely transformations \cite{FirstStereo1976, barnard1982computational, calcOpticalFlowFarneback, hirschmuller2008stereo}.
These algorithms were based on geometric assumptions about the physical world,
but faced practical limitations when confronted with hard to quantify
optical artefacts, like lensing, reflections, specular lighting,
refractions or sensor degradation.
A big leap was achieved with the advent of convolutional neural networks (CNN), which
have proven to be extremely effective in image recognition and understanding \cite{CNN1, CNN3, CNN2}.
Given enough training data, these algorithms were proven to excel in both, accuracy and speed for
a wide range of tasks, including image registration. Additionally, neural networks are the only technology
that has the potential to expand into more abstract interpretations, where, for example, correspondes are formed
based function in relatably designed objects \cite{lai2021functionalcorrespondenceproblem}.

Unfortunately CNNs do not solve the correspondence problem directly from input images. They are not suited to measure
pixel-precise distances across large regions, because they capture increasingly abstract features in layers that
would be high enough to cover a large enough input window. First applications of CNNs were therefore
limited to the assessment of matching costs, i.e. how well the feature seen in one image matches its
presumed twin in the other image \cite{CNNCost2, CNNCost1, CNNCost3}.
A computationally intensive application of CNNs is then to pick the best
match from a 3- or 4-dimensional tensor containing the image in all possible displacements, known as cost volume or cost cube
\cite{xue2022stereomatchingcostvolume}. A pyramidal network layout made it possible to reduce
the size of the volume based coarse estimates on scaled versions of the input images \cite{PyramidStereoMatchingNetwork,PyramidNetwork2016}.
These methods have proven very effective in applications of stereo vision and scene flow
\cite{OpticalFlow2015, li2021revisitingstereodepthestimation, kim2024unitt}.
Top performing algorithms use a combination of various methods to infer object boundaries from
monocular vision cues, combined with stereo vision and established properties of the geometric world
\cite{wen2025foundationstereozeroshotstereomatching, SceneFlowEstimation, OpticalFlow}.
The complexity and the inherent ties to 3D scenes make these algorithm difficult to adapt to simpler scenarios,
where little training data is available and correspondence detection must be rebuilt from scratch.

This article introduces a recursive formulation to an end-to-end trainable algorithm that aimes for simplicity and generality.
It is based on a single network that is trained from ground truth with few assumptions about the task.
Previous works have utilised coarse-to-fine inferences with 2 \cite{CascadeResidualLearning}
and 4 stages \cite{li2022practicalstereomatchingcascaded} of corrections without building a cost volume.
This algorithm reformulates these approaches in a recursive fashion without limits to the number of stages.
Its implementation spans only few lines of Python code, forming a tiny wrapper around a single CNN that is trained to
work on all levels.
The algorithm was first applied to compare relevant changes in screenshots in the presence of distorting, but valid layout changes \cite{infoq}.
This article explores a practical application to stereoscopy.
For this experiment a single CNN with 19x15 input pixels was trained. The results are not
always maximally accurate, but tiny in computational footprint.
The benefit of this approach is the simplicity of the code and the ease of training.

\section{Problem statement}\label{problem}

Two images are given as functions $f_1, f_2 : \mathbb{R}^n \to \mathbb{R}^m$, that describe an $n$-dimensional image
with $m$ channels. We assume that $f_1$ and $f_2$ are observations of the same shared physical space.
They could represent 2D images measured in 3 colors, but the derivation of the following algorithm is more general. Error bounds and
algorithmic steps could easily be applied to video sequences, volumetric measurements and other types of data that
may be pertinent to the task.
We assume that any transformation might have happened between both measurements in $f_1$ and $f_2$.
The task of the correspondence problem is to capture the transport related part of that transformation.
If a physical feature measured at position $x$ in $f_1$ and it has moved by a distance of $d(x)$
then it would be measurable at position $x - d(x)$ in the second image $f_2$.
The vector $d$, also known as disparity, describes the motion from image 2 to 1, or
as it is interpreted in stereo vision, the apparent motion caused a camera change from image 1 to 2.
\begin{equation}\label{eqn:correspondence}
    f_1\big(x) \cong  f_2\big(x - d(x)\big)
\end{equation}

\subsection{Objective}
To formalize the objective we are looking for a displacement field $d: \mathbb{R}^n \to \mathbb{R}^n$,
such that it optimally maps between the coordinates in the two given images.
We define the translation operator $T$ such that it applies the transformation described by $d$ to an image $f$.
\begin{equation}\label{eqn:translation}
    T\big(f, d\big)(x) := f\big(x - d(x)\big)
\end{equation}

When applying a candidate transformation $d$ to image $f_2$ then it should be aligned with $f_1$. The quality of the alignment is
known as the matching cost. It measures how well the $m$-dimensional image features match at each position.
Mathematically, this is expressed by a suitable norm in the function space.
Hence, we are looking for a displacement that minimizes the difference between the first and the transformed second image.
\begin{equation}
    d = \arg \min_{d'} \big\| f_1 - T(f_2, d') \big\|
\end{equation}

Inserting into the correspondence condition (\ref{eqn:correspondence}) this means that both images are now aligned for
all points.
\begin{equation}
    f_1 \cong T(f_2, d)
\end{equation}

\subsection{Constraints}
\begin{enumerate}
\item \textbf{Velocity bound}:
The displacement is bounded by a maximum value $M$.
This reflects the inherent finite nature of practical applications.
\begin{equation}
\| d(x) \| < M
\end{equation}

\item \textbf{Continuity bound}:
This bound $\lambda$ limits the disparity difference between neighboring points.
It forbids sudden jumps in the displacement field and limits how sharp the disparity can be resolved for nearby objects
moving at different speeds. Jumps that are larger than $\lambda$ will not be resolved by the following algorithm and remain blurry.
\begin{equation}\label{eqn:non-intersection}
    \big\| \nabla d(x) \big\| < \lambda
\end{equation}

\end{enumerate}

\section{The algorithm}

The following algorithm is a recursive scheme that leverages a neural network that estimates the displacement field of up to a small distance of $\mu$ and
turns it into an algorithm that detects large displacements up to $M$. The slope bound $\lambda$, i.e. the resolution at which objects can be resolved is
not changed by the recusive scheme. Real optical flows often have discontinuous boundaries between
occluding objects that move at different speeds. Such boundaries cannot be resolved precisely with this algorithm, but the values away from boundaries can.
If the speed difference is $\Delta v$ then the blurry region must have a width of $\|\Delta v\| / \lambda$. This unchartered region is proportional to the amount of new
stuff becomming visible from image to image. For $\lambda>1$ the error is small, when compared to the part that needs to be solved by pattern segmentation
(or monocular vision) anyways. The presented algorithm only solves the part of visual correspondence and does not make any assumptions about
the geometric properties of the domain. Hence, it is generic and simple, but not always precise.

\subsection{Schematic view of the algorithm}

\begin{figure}[h]
\centering
\includegraphics[width=1.0\textwidth]{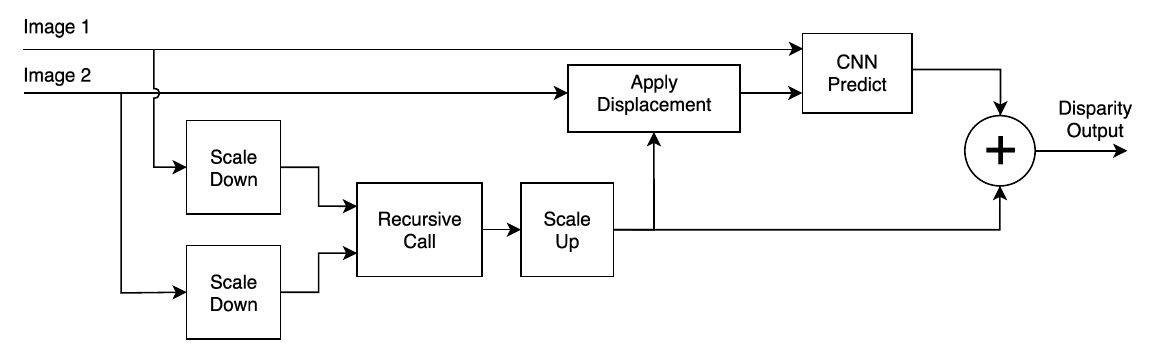}
\caption{Data flow in the recursive scheme. The algorithm works on
    scaled down versions of the input images and then corrects the result.}
\label{fig:recursion}
\end{figure}

The flow chart in figure \ref{fig:recursion} depicts the fundamental idea of the recursive algorithm.
It is based on a multi-scale approach,
where a coarse estimate is refined by a subsequent correction step.
The coarse estimate is generated by scaling the input images down and then calling
the algorithm recursively on the lower resolution. The result obtained on the smaller scale is then
scaled back to the original size. Based on this approximate disparity, one of the input
images is transformed to reverse the effect of the assumed displacement, as defined in equation (\ref{eqn:translation}).
If the coarse estimate was perfect and there was no noise, both images, one transformed and one untransformed,
would then be identical.
Given the limitations of finite resolutions the images will differ and may require an additional push.
To estimate the required correction a convolutional neural network (CNN) is trained to predict
small disparities between image patches, effectively operating as a sliding window that corrects remaining disparities.
The network is oblivious to recursion depth and its relative position. It can be trained from one source of truths.
The final result is the sum of the coarse estimate and the correction.

The recursion has to end at some point. When the image is getting to small to make meaningful
inferrences, the algorithm has to assume a displacement of zero.
This means that no displacment is applied to the input images, which, at this level of recursion,
represent the original input after multiple scaling operations. They are fed into the CNN without additional transformations.
From then on, the CNN works to correct its own outputs.

\subsection{Accuracy Considerations}\label{algorithm}

We start our recursive construction of the final algorithm by assuming that we
already solved the problem for small displacements up to some value of $\mu$. This
base algorithm is called $A^\mu$. A convolutional neural network
has no problem detecting correspondences over a small number of $\mu$ pixels.
\begin{equation}
    A^\mu(f_1, f_2) \approx d\quad\mbox{ for } \|d(x)\|< \mu
\end{equation}

The error of that initial algorithm shall be bounded by $\mu/2$.
\begin{equation}\label{eqn:error_mu}
    \| A^\mu(f_1, f_2)(x) - d(x) \| < \frac{\mu}{2}
\end{equation}

In a next step we want to extend the solution for displacements up to distances of $2\mu$
in a repeatable scheme that will eventually bring us to the global bound of $M$.
For this we introduce a scaling operator $S$ and its inverse $S^{-1}$ that can zoom in and out of the arguments.
\begin{eqnarray}
    S f(x) &:=& f(2 x) \nonumber \\
     S^{-1} f(x) &:=& f(\frac{1}{2}x)
\end{eqnarray}

By applying this operator we can scale down all distances and use the algorthm $A^\mu$ to find
correspondences on a view of the functions where distances of $2\mu$ are shortened to $\mu$.
This operation maintains the continuity bound (\ref{eqn:non-intersection}), becauuse the slope increase
caused by S is compensated by the shortening of path length.
The resulting displacments must then be scaled up again and the whole precedure is reversed.
Now we are working on disparity fields and not images. Therefore we must explicitely multiply the result by two and
convert the measured distances back to the original scale.
The preliminary algorithm $\tilde{A}^{2\mu}$ calculates this procedure.
\begin{equation}
    \tilde{A}^{2\mu} := 2 S^{-1} A^\mu (Sf_1, Sf_2)
\end{equation}

We had to multiply the displacement result by two, because scaling the function with
$S^{-1}$ also doubles all distances between corresponding points. Unfortunately,
this also increases all errors by a factor for two, raising the uncertainty from
its original value  $\mu/2$ in (\ref{eqn:error_mu}) to $\mu$ in (\ref{eqn:error_mu2}).
\begin{equation}\label{eqn:error_mu2}
    \| \tilde{A}^{2\mu}(f_1, f_2)(x) - d^\star(x) \| < \mu
\end{equation}

Having established the approximate algorithm $\tilde{A}^{2\mu}$ we have gained the ability to capture the correspondence
of points twice as distant as the initial one. Now we have to fix the lost accuracy in this construction.
We apply the translation operator (\ref{eqn:translation}) to the the second argument and reduce
the remaining disparity and move corresponding features closer together.
If the algorithm was perfect then the functions $f_1$ and $T(f_2, d)$ would be
perfectly aligned with no remaining disparity. However, with an error of up to $\mu$,
in the worst case a dispariy of $2\mu$ is measured as $2\mu \pm \mu$. In this case the translation operator $T$
moves the corresponding points into the right direction, but still leaves behind a gap of maximum length $\mu$.
In other words, the disparity maximum dropped from $2\mu$ to $\mu$. Coincidentally, this falls nicely into the range of
our original algorithm $A^{\mu}$, which has the ability to detect the remaining displacement
with an error of $\mu/2$. The resulting algorithm, called $A^{2\mu}$, adds the estimates on a coarse scale to
the correction term estimated on the fine scale. This increases the search range,
without compromising on accuracy.
\begin{equation}
    A^{2\mu}(f_1, f_2) := \tilde{A}^{2\mu}(f_1, f_2) + A^{\mu}\left(
        f_1, T\left(
            f_2,
            \tilde{A}^{2\mu}(f_1, f_2)
        \right)
    \right)
\end{equation}

With this construction we can now recursively create a disparity tracker for large displacements
without loss of accuracy. One only needs an initial algorithm that works up to a positive $\mu>0$ and
ensures an error bound below $\mu/2$. Any arbitrary distance bound $M$ can be achieved with finite
iterations. Finding correspondences over arbitrarily short distances is a problem,
for which many solutions exist. Most easily this is solved with
a neural network that either sees a massively scaled down version of the input data,
or it sees a tiny fractions of the input where most of the displacement has already been corrected
for.

\section{Implementation for Image Data}\label{sec:implementation}

This sections shows a 2D implementation of the algorithm in runnable Python code. It
relies on a few helper functions as outlined below.
The main function {\tt get\_\allowbreak correspondence\_\allowbreak map} implements the recursive scheme $A^M$ and
computes the disparity over two images {\tt img1} and {\tt img2}.
It assumes there is already a pretrained convolutional neural network to predict
small displacements.
\lstset{language=Python}
\begin{lstlisting}
# Returns a tensor d, such that for every x, y:
# img1[y, x] corresponds to img2[y - d[y,x,1], x - d[y,x,0]]
def get_correspondence_map(img1, img2):
  assert_equal(img1.shape, img2.shape)
  if window_size <= img1.shape:
    # First estimate with a recursive call.
    d1 = 2 * resize(
      get_correspondence_map(
        resize(img1, 0.5),
        resize(img2, 0.5)), 2)
    # Calling CNN to predict residual displacements.
    d2 = cnn_predict(img1, transform(img2, d1))
    return d1 + d2
  else:
    # If the image is too small, assume zero displacement.
    return zeros((img1.height, img1.width, 2))
\end{lstlisting}

After the first iteration the neural network is not applied to scaled down original images, but to
precorrected images. This is fundamentally the same task, but might face some unexpected artefacts that do
not occur otherwise, such as e.g. reverse directions when the first estimate was too big.
Hence, this is a bootstrapping problem. Without running the algorithm we can't have training data for precorrected input.
Without a trained network we can't run this algorithm.

\subsection{Helper Functions}

\subsubsection*{\tt window\_size}
The window size is minimal image size for which the
CNN can make predictions. The {\tt <=} operator should yield true accordingly.

\subsubsection*{\tt resize(..., 0.5)}
The first subroutine to be executed is the scale down operation, $S$.
Here the image size is reduced by a factor of two. All image libraries provide such function.
Usually, this squeezes the information of neighboring pixels into one, by averaging their
values.

\subsubsection*{\tt resize(..., 2)}
With an argument of 2 the function implements $S^{-1}$, i.e. the inverse of the previous scale.
It brings an image back to the original size. Additional values for the new interjected pixels must be
interpolated linearly. An improvement could be made here with a custom upscaling method that considers the
image's color information to decide how to interpolate.

\subsubsection*{\tt transform(img, d)}
This is the implementation of the translation operator $T$.
The computer vision library OpenCV provides such a function out of the box as {\tt remap}, but requires
absolute coordinates instead of the relative ones that can be predicted from a translation
invariant window.
For absolute positions a NumPy {\tt meshgrid} must be added. The following is Python code for this operation.

\begin{lstlisting}
def transform(img, d):
    x, y = np.meshgrid(np.arange(img.shape[1]),
                       np.arange(img.shape[0]))
    return cv2.remap(img,
                     np.float32(x - d[:,:,0]),
                     np.float32(y - d[:,:,1]),
                     interpolation=cv2.INTER_AREA,
                     borderMode=cv2.BORDER_REPLICATE)
\end{lstlisting}

\subsubsection*{\tt cnn\_predict(img1, img2)}
This subfunction computes a predicted disparity from a neural network.
It takes two input images with a matching number color channels. It outputs the disparity as an image of the
same size with 2 channels, one for $x$ and one for $y$ components. It must be a convolutional style
network that can work on arbitrary image sizes. CNNs operate on sliding windows where each output
pixel is predicted from a number of surrounding inputs in a sliding window.
Section \ref{sec:training} discusses how to train such a network.

\subsection{Training}\label{sec:training}
The neural network can be trained from physically measured data or from synthetically created images.
The easiest way of getting ground truth data is applying
random distortions to images. This requires that we can randomly create
distortions that are pertinent to our problem domain.
For many real world applications it is difficult to foresee the types of artefacts that can appear in input data.
Using realistic footage is the only option when such effects cannot be synthesized. Luckily some high precision data
sources for 3D vision are available online \cite{scharstein2003}. The CNN for this algorithm can
be built extremely efficiently without the need for large amounts of training data.
In the first, and coarsest, application of the network, it is applied to the input image on a maximally scaled down version.
Then, every subsequent inference is based on an image that is rectified based on previous predictions.
This has the nice side effect that this algorithm creates its own synthetic training data.
Every time the weights are updated, the predicted distortions change and new images are generated
for the next inference. However, with an initial set of random weights this produces misfitted training data
after the first recursion. To avoid this, training must start from tiny images, where
the algorithm makes no recursive calls. In subsequent training rounds, the number of recursions can be increased.
For each call {\tt cnn\_predict(img1, transform(img2, d1))} we can extract the expected training truth as
{\tt resize(ground\_truth, d1.shape) - d1}. This evaluates the required offset, i.e. the remaining mismatch between
the true value and the previous prediction. To stabilize convergence a Gaussian blur can smoothen the training goal.
Large values must be excluded from loss calculation, because predictions are not possible if the
corresponding features are barely visible within the same input window.

\section{Application to Stereopsis}

Stereopsis is one of the most common applications of the correspondence problem.
In this setup, image 1 is taken from the left camera and image 2 from the right. The disparity describes
the apparent leftward displacement of items in the second image.
The produced disparities are inversely proportional to distances from the left camera. This
is an important information for autonomous navigation and near distant spacial sensing.
The problem is significantly simplified. Both cameras are guaranteed to be epipolar, i.e. shifted along a known axis.
Only the $x$-component of the disparity needs to be determined. We can assume $y$ to
always be zero.

The results presented in this section were created with a neural network of
input size 19 by 15 pixels by 6 channels of RGB values
for both images. The output is one channel for $d$ and one output pixel for each input pixel.
To ensure equal input and
output size, inputs are padded before each evaluation. The trainable weight count was around 550k
and ran for a couple of hours on an M1 MacBook Pro. The simplicity of the network and
the speed of convergence make this algorithm ideal as a base algorithm for more select
use cases.

As ground truth 74 image pairs were taken from Middlebury data set for stereopsis \cite{scharstein2014high, scharstein2002taxonomy, hirschmuller2007evaluation}.
The images were ideal for training, very sharp and showed almost no glares or refractions.
To create additional variety, images were mirrored and hues were randomized.
Still, the number of images is very low when compared to typical training sets for CNNs.
The accuracy of the obtained results demonstrates the effectiveness of self-synthetization
of training data. The reusability of one weight set on all levels of details is crucial. It
prevents overfitting and enables cross learning between distant and close
objects.

\subsection{The CNN Model}

Table \ref{tab:cnn_model} shows the architecture of the CNN model.
There has been little hyperparameter tuning and no destillation, leaving lots of room for
improvement. Especially, when considering more specific use cases one could almost
definitely adapt the network for speed and/or accuracy. The benefit of this approach is
its simplicity and adaptability.

\begin{table}[htbp]
\centering
\caption{CNN model summary. The kernel size is calculated for one ouput pixel.
    Inputs need to be padded to reach the desired output shape.}
\label{tab:cnn_model}
\begin{tabular}{cccccc}
\textbf{Layer} & \textbf{Type} & \textbf{Config} & \textbf{Output Shape} & \textbf{Parameters} & \textbf{Activation} \\
\hline
0 & Input & - & (15, 19, 6) & 0 & - \\
1 & Conv2D & (3,3) & (13, 17, 12) & 660 & relu \\
2 & Conv2D & (3,3) & (11, 15, 24) & 2,616 & relu \\
3 & Conv2D & (3,3) & (9, 13, 32) & 6,944 & relu \\
4 & Dropout & 0.1 & (9, 13, 32) & 0 & - \\
5 & Conv2D & (3,3) & (7, 11, 46) & 13,294 & relu \\
6 & Conv2D & (3,3) & (5, 9, 72) & 29,880 & relu \\
7 & Dropout & 0.1 & (5, 9, 72) & 0 & - \\
8 & Conv2D & (1,3) & (5, 7, 100) & 21,700 & relu \\
9 & Conv2D & (3,3) & (3, 5, 200) & 180,200 & relu \\
10 & Conv2D & (1,1) & (3, 5, 200) & 40,200 & relu \\
11 & Conv2D & (3,3) & (1, 3, 128) & 230,528 & relu \\
12 & Conv2D & (1,3) & (1, 1, 64) & 24,640 & relu \\
13 & Conv2D & (1,1) & (1, 1, 32) & 2,080 & relu \\
14 & Conv2D & (1,1) & (1, 1, 1) & 33 & linear \\
\hline
\multicolumn{4}{r}{\textbf{Total Parameters:}} & \textbf{552,795} & \\
\end{tabular}
\end{table}

\subsection{Validation Data}
\begin{figure}[h]
\centering
\includegraphics[width=1.0\textwidth]{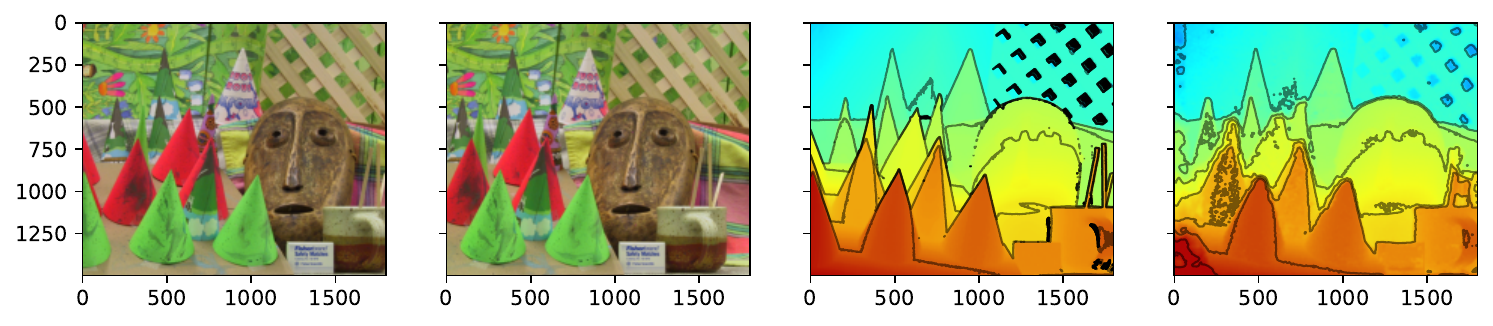}
\includegraphics[width=1.0\textwidth]{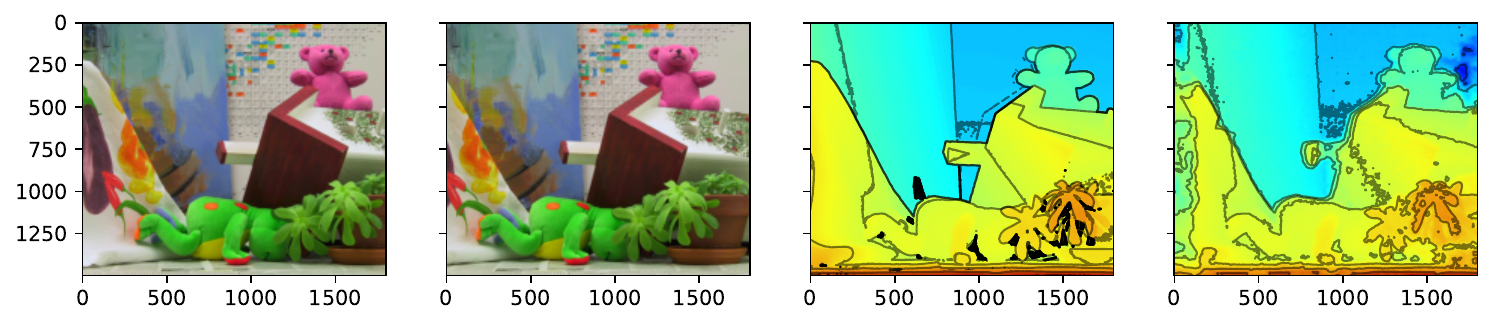}
\caption{Left and right camera view on two scenes. The ground truth was not used during training. Results in column four.}
\label{fig_cones}
\end{figure}

Two images were withheld from the training procedure, but might have inadvertently been used to
make the decision to finalize this analysis and start publishing. The images were originally
published in \cite{scharstein2003} and have since been established as a common benchmark for
stereopsis algorithms. The objects in this image have a rich texture and no glares, ideal
for stereo vision applications. Figure \ref{fig_cones} shows the two images, the ground thruth and the prediction.

Table \ref{tab_bad} shows that error rates that are on par with other algorithms in that field.
The number of pixels with deviations between predicted and actual disparity is around 10\%.
Coincidentally, this is also the number of pixels that are occluded in either of the two views.
The only way to get better results is to combine this algorithm with monocular inference, such as shadows
and an understanding of common 3D shapes. The percentage of perfect inferences is close to 90\%, which
emphasises the fact that this algorithm can be absolutely precise even on high disparities of 200 pixels in
an image of 1800x1500, if depicted objects are sharp and textured.
The small input window does not inherently limit the precision,
although it obviously lacks a deeper understanding of the 3D world.
\begin{table}[ht]
\caption{Percentage of bad pixels for the two data sets.}
\label{tab_bad}%
\begin{tabular}{lrrrrr}
Data set & $P(e>1)$ & $P(e>2)$ & $P(e>5)$ & $\max(e)$ & partial occlusion \\
\midrule
Cones & 13.0\% & 10.9\% & 8.5\% & 102px & 16.3\%\\
Teddy & 11.7\% & 8.5\% & 5.8\% & 97px & 12.8\%\\
\botrule
\end{tabular}
\end{table}

\subsection{Test Data}

\begin{figure}[ht]
\centering
\includegraphics[width=1.0\textwidth]{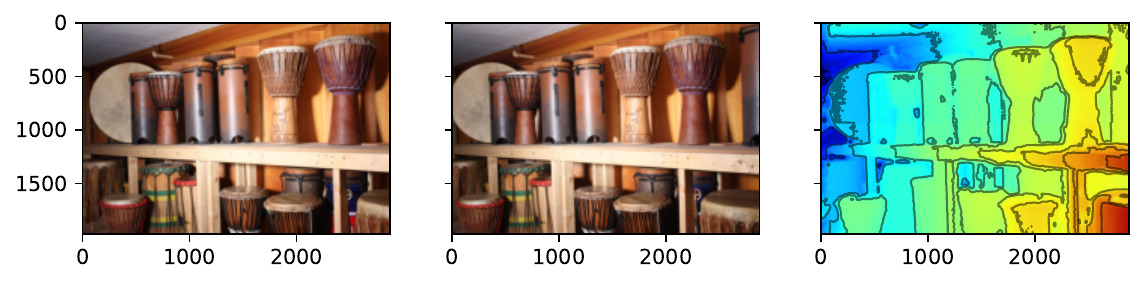}
\includegraphics[width=1.0\textwidth]{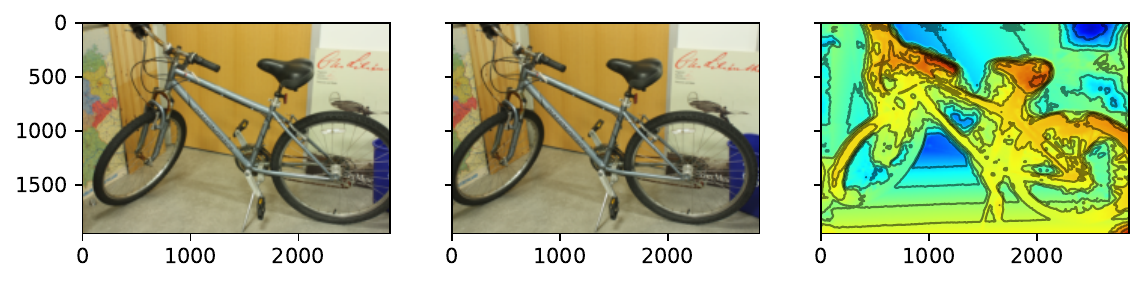}
\includegraphics[width=1.0\textwidth]{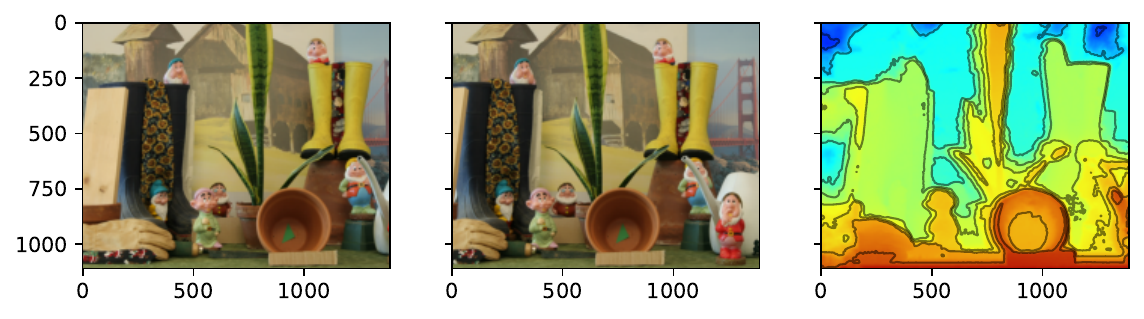}
\includegraphics[width=1.0\textwidth]{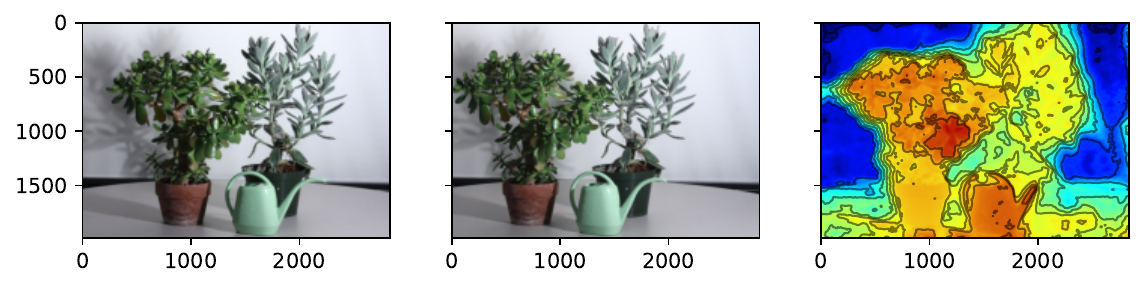}
\includegraphics[width=1.0\textwidth]{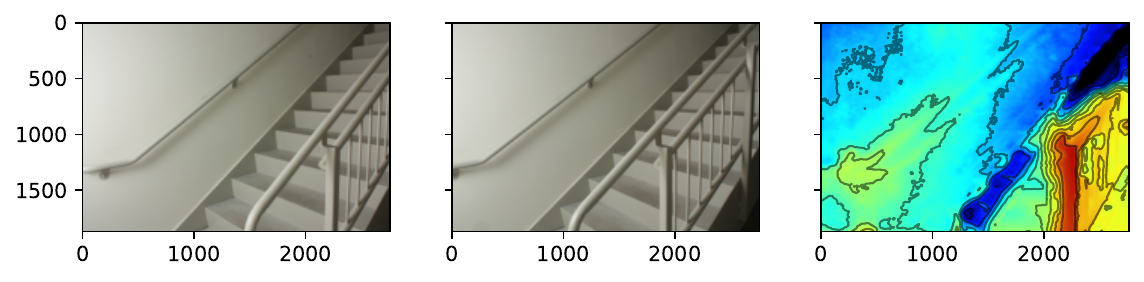}
\caption{Left and right camera view, followed by the inferred disparity. The algorithm measures the disparities
         of extended landmarks precisely, but performs poorly on discontinuous geometries where monocular inference is required.}
\label{fig:unannotated}
\end{figure}

Figure \ref{fig:unannotated} shows a cross section of the results from unannotated test images. The ground truth
for these images were withheld by the publishers. As humans we can nontheless qualitatively assess the accuracy of the results.
We are so intimately familiar with the depicted objects that we just know how they would shape out in 3D space,
even from a single image.

When judging the results it should be considered, that the network was trained on limited data with few hours of training time
on stock hardware. It relies soley on stereo vision to infer depth and did not see enough training samples to understand 3D geometries.
Hence, it underperforms on sharp edges and large regions without texture.
The worst result is shown in the final row, the staircase. There are no colors,
no texture, a complex geometry and worst of all, there was no remotely similar scene in the training set.
Nevertheless, depth inference looks quite promising overall, especially when considering the simplicity
and the make shift nature of the entire approach.

\section{Conclusion}\label{sec13}

This article introduced a recursive algorithm for solving the image registration problem,
demonstrated through an application to stereopsis. The key contribution is a recursive formulation
of the algorithm that works with one simple neural network. It can be implemented in just a dozen lines of code while
maintaining competitive accuracy in the regions for which it was trained.
It was shown to produce pixel accurate results for continuous regions, but remained blurry at discontinuous boundaries,
as they are produced by occluding projections. The preferred methods to resolve such edge cases is very much domain specific,
as the projection artefacts seen in fields like medicine, astronomy and 3D projections are all different and demand bespoke solutions.
The benefit of the presented approach is its algorithmic beauty and simplicity, allowing it to be adaptable and serve as a basis
for further exploration.

The recursive formulation offers several practical advantages: it requires minimal training data
(demonstrated with only 74 image pairs), enables efficient training through self-generated synthetic data,
and provides a lightweight solution suitable for resource-constrained applications. The algorithm achieved
error rates comparable to established methods while using a compact (undistilled) 550k parameter network from a
19x15 input window.

Future work should focus on:
(1) training the scale down operator $S$ to maintain relevant information, possibly projecting into more than just 3 color channels,
(2) training the scale up operator $S-1$ to utilize color information instead of plain linear interpolation,
(2) preprocessing the input image to $m$-dimensional feature embeddings instead of the original 3 color channels
(4) integration with explicit monocular depth cues for handling occlusioned regions, and (5)
extension to additional sensory inputs, such past scenes with data from motion sensors, additional cameras or lidar scanners.

\backmatter

\section*{Declarations}

Stefan Dirnstorfer is a cofounder of testup.io, a venture that markets solutions for visual software testing.

\bibliography{article}% common bib file

\end{document}